\documentclass[10pt,twocolumn,letterpaper]{article}

\usepackage{iccv}
\usepackage{times}
\usepackage{epsfig}
\usepackage{graphicx}
\usepackage{amsmath}
\usepackage{amssymb}
\usepackage{threeparttable}
\usepackage{booktabs}
\usepackage{multirow}
\usepackage{authblk}

\usepackage[pagebackref=true,breaklinks=true,letterpaper=true,colorlinks,bookmarks=false]{hyperref}
\iccvfinalcopy 
\pdfoutput=1 
\def\iccvPaperID{736} 

\ificcvfinal\fi
\begin{document}
\language0
\lefthyphenmin=2
\righthyphenmin=3

\newcommand{\lstm}{\textrm{L}^\textrm{2}\textrm{STM}}
\newcommand{\FSTCN}{\textrm{F}_{\textrm{ST}}\textrm{CN}}
\title{Lattice Long Short-Term Memory for Human Action Recognition}

\author[1,2]{Lin Sun}
\author[3]{Kui Jia}
\author[2]{Kevin Chen}
\author[1]{Dit Yan Yeung}
\author[1]{Bertram E. Shi}
\author[2]{Silvio Savarese}
\affil[1]{The Hong Kong University of Science and Technology}
\affil[2]{Stanford University} \affil[3]{South China University of Technology}
\maketitle
\begin{abstract}
Human actions captured in video sequences are three-dimensional signals characterizing visual appearance and motion dynamics. To learn action patterns, existing methods adopt Convolutional and/or Recurrent Neural Networks (CNNs and RNNs). CNN based methods are effective in learning spatial appearances, but are limited in modeling long-term motion dynamics. RNNs, especially Long Short-Term Memory (LSTM), are able to learn temporal motion dynamics. However, naively applying RNNs to video sequences in a convolutional manner implicitly assumes that motions in videos are stationary across different spatial locations. This assumption is valid for short-term motions but invalid when the duration of the motion is long.

In this work, we propose Lattice-LSTM ($\lstm$), which extends LSTM by learning independent hidden state transitions of memory cells for individual spatial locations. This method effectively enhances the ability to model dynamics across time and addresses the non-stationary issue of long-term motion dynamics without significantly increasing the model complexity. Additionally, we introduce a novel multi-modal training procedure for training our network. Unlike traditional two-stream architectures which use RGB and optical flow information as input, our two-stream model leverages both modalities to jointly train both input gates and both forget gates in the network rather than treating the two streams as separate entities with no information about the other. We apply this end-to-end system to benchmark datasets (UCF-101 and HMDB-51) of human action recognition. Experiments show that on both datasets, our proposed method outperforms all existing ones that are based on LSTM and/or CNNs of similar model complexities.
\vspace{-1em}
\end{abstract}

\section{Introduction}

\begin{figure}[htp]
  \centering
  \centerline{\includegraphics[width=8cm, height=5cm]{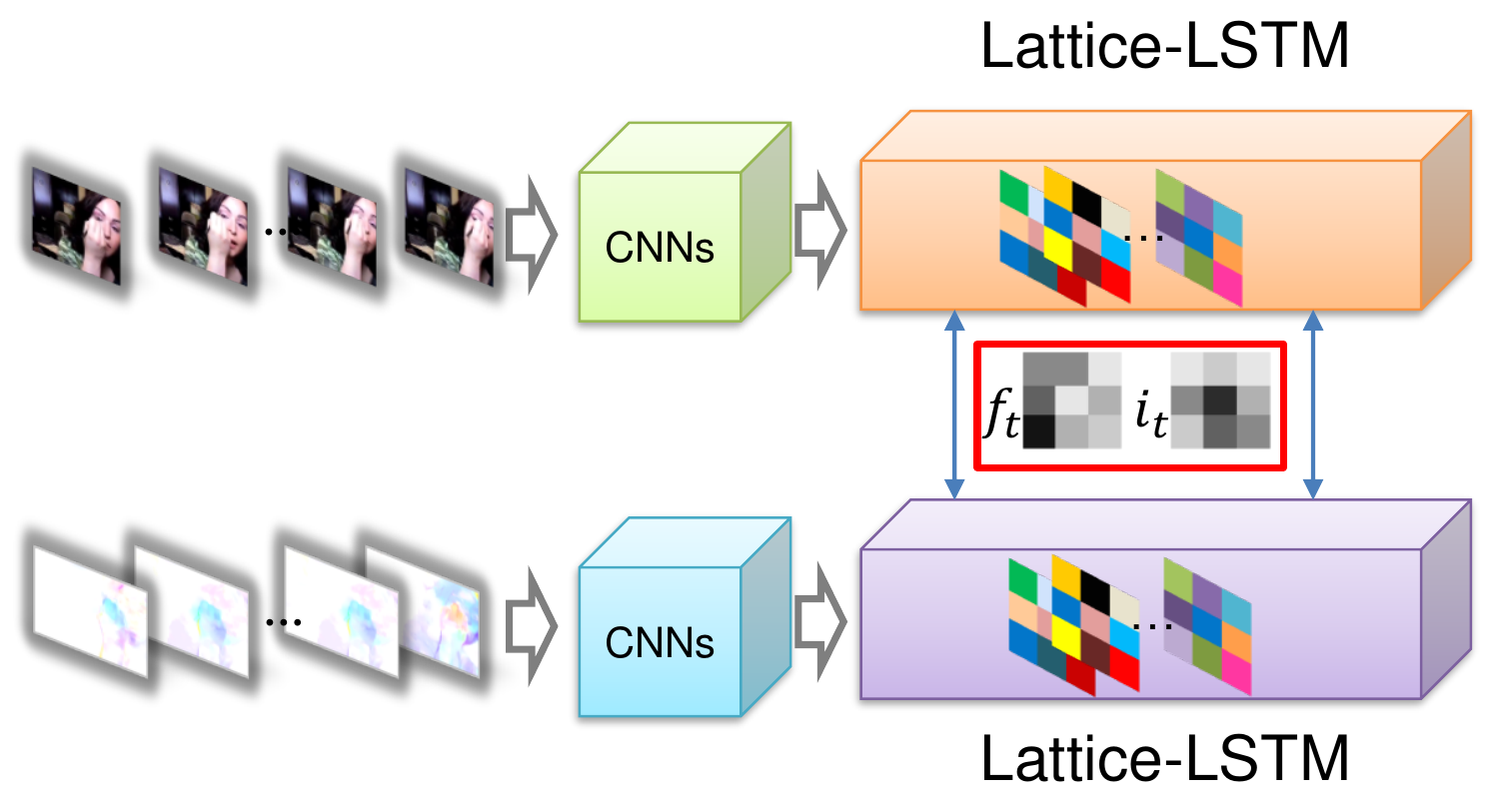}}
  \caption{The mechanism of Lattice-LSTM ($\lstm$). $f_t$ and $i_t$ represent the forget and input gate patterns at time $t$, respectively. Two technical contributions are illustrated here: Firstly, more motions, particularly complex ones, can be depicted using a spatially varying local superposition within the memory cell. Secondly, input and forget gates are multi-modal trained using RGB and corresponding optical flow. They will serve as the motion attention mask controlling the entering/leaving dynamics from the memory cell. Even more motion dynamics are generated from the memory cell, but only the useful ones are preserved for representation. } \label{fig:llstm}
\end{figure}

Video based human action recognition is of importance for many applications, including surveillance\cite{vid_surveillance}, abnormal activity detection \cite{vid_unusual}, human computer interaction (HCI) \cite{vid_HCI} and so forth. Unlike recognition in static images, human action recognition relies on motion dynamics in addition to visual appearance. Designs of many human action recognition datasets \cite{HMDB-51, UCF-101} are based on this observation.

Following the success of Convolutional Neural Networks (CNNs) on computer vision tasks such as image classification, researchers have achieved remarkable performance on public human action recognition datasets. Deep learning has been shown to have an extraordinary ability to extract powerful representations. In order to handle videos, CNNs have been applied to videos on a per-frame basis, followed by simple or strategic pooling across the temporal domain. However, CNNs only consider appearance features frame by frame without any consideration of interactions across the temporal domain. Meanwhile, the traditional Long Short-Term Memory (LSTM) with matrix multiplications does not take spatial layout into consideration. ConvLSTM \cite{xingjian2015convolutional} and VideoLSTM \cite{li16videolstm} address this limitation by applying convolutional operations on sequences of images or feature maps. These models tend to work well for actions with little movement, which we refer to as stationary motions. However, the limited modeling capacity of these architectures causes the network to struggle with long-term and complex motions, as verified by our experiments. An accurate action recognition should (1) have a high capacity for learning and capturing as many motion dynamics as possible, and (2) when an action appears in sequential images, the neurons should properly decide what kind of spatio-temporal dynamics should be encoded into the memory for distinguishing actions.

In this paper, we propose a novel two-stream LSTM architecture named Lattice-LSTM ($\lstm$). $\lstm$ extends LSTM by learning independent memory cell transitions for individual spatial locations with control gates that are shared between RGB and optical flow streams. It can more effectively address the non-stationary issue of long-term motion dynamics without significantly increasing the model complexity. At the same time, the control gates are trained using multiple modalities, RGB images and optical flow, in order to better control the information entering/leaving the memory cell. We summarize the main contributions of this paper as follows:
\begin{itemize}
\item Our derivation of the recurrent relation confirms our assumption that the current LSTM architecture, including LSTM, ConvLSTM and VideoLSTM, cannot accurately model the long-term and complex motions within videos. A novel LSTM, called Lattice-LSTM ($\lstm$), that can greatly enhance the capacity of the memory cell to learn motion dynamics is proposed. Extensive experiments have been conducted to verify the effectiveness of the proposed $\lstm$.

\item To leverage both RGB and the corresponding optical flow information, the two input gates and the two forget gates of our two-stream $\lstm$ are shared and trained by both modalities, which achieves better control of the dynamics passing through the memory cell.

\item We propose a new sampling strategy to enable RNNs to learn long- and short-term temporal information from randomly extracted clips. Compared with consecutive frame sampling and sampling with strides, our sampling method provides data augmentation along the temporal domain, which helps the learning of recurrent dynamics.

\item We present a novel architecture to enhance, encode and decide motion dynamics for action recognition. Experiments on the benchmark UCF-101 and HMDB-51 datasets show that our proposed methods outperform all existing ones that are based on LSTM and/or CNNs of similar model complexities.
\end{itemize}
\vspace{-1em}


\section{Related Work}

Following the great success of CNNs for image classification \cite{AlexNet}, image segmentation \cite{dl-segmentation}, image retrieval \cite{dl-retrieval} and other computer vision tasks, deep learning algorithms have been used in video based human action recognition as well. Karpathy et al. \cite{vid_cnn} directly apply CNNs to multiple frames in each sequence and obtain the temporal relations by pooling using single, late, early and slow fusion; however, the results of this scheme are just marginally better than those of a single frame baseline, indicating that motion features are difficult to obtain by simply and directly pooling spatial features from CNNs.

In light of this, Convolutional 3D (C3D) is proposed in \cite{C3D} to learn 3D convolution kernels in both space and time based on a straightforward extension of the established 2D CNNs. However, when filtering the video clips using 3D kernels, C3D only covers a short range of the sequence. Simonyan et al. \cite{TwoStream} incorporate motion information by training another neural network on optical flow \cite{optflow}. Taking advantage of the appearance and flow features, the accuracy of action recognition is significantly boosted, even by simply fusing probability scores. Since optical flow contains only short-term motion information, adding it does not enable CNNs to learn long-term motion transitions. 

Several attempts have been made to obtain a better combination of appearance and motion in order to improve recognition accuracy. Lin et al. \cite{fstcn} try to extract the spatial-temporal features using a sequential procedure, namely, 2D spatial information extraction followed by 1D temporal information extraction. This end-to-end system considers the short (frame difference) and long (RGB frames with strides) motion patterns and achieves good performance. Feichtenhofer et al. \cite{Feichtenhofer16fusion} study a number of ways of fusing CNN towers both spatially and temporally in order to take advantage of this spatio-temporal information from the appearance and optical flow networks. They propose a novel architecture \cite{Feichtenhofer16resinet} generalizing residual networks (ResNets) \cite{he2016resnet} to the spatio-temporal domain as well. However, a CNN based method cannot accurately model the dynamics by simply averaging the scores across the time domain, even if the appearance features already achieve remarkable performance on other computer vision tasks.

In order to model the dynamics between frames, recurrent neural networks (RNNs), particularly long short-term memory (LSTM), have been considered for video based human action recognition. LSTM units, first proposed in \cite{lstm}, are recurrent modules which have the capability of learning long-term dependencies using a hidden state augmented with nonlinear mechanisms to allow the state to propagate without modification. LSTMs use multiplicative gates to control access to the error signal propagating through the networks, alleviating the short memory in RNNs \cite{rnn_timing}. \cite{xingjian2015convolutional} extends LSTM to ConvLSTM, which considers the neighboring pixels' relations in the spatial domain. ConvLSTM can learn spatial patterns along the temporal domain.

LSTMs have achieved remarkable performance on language modeling \cite{lstm-app}, but their performance on video action recognition still lags. \cite{action-lstm}, \cite{two-lstm} and \cite{LRCN} propose LSTMs that explicitly model short snippets of ConvNet activations. Ng et al. \cite{two-lstm} demonstrate that two-stream LSTMs outperform improved dense trajectories (iDT) \cite{iDT} and two-stream CNNs \cite{TwoStream}, although they need to pre-train their architecture on one million sports videos. Srivastava et al. \cite{Srivastava_unsupervisedlearning} propose an interesting LSTM based unsupervised training method using an encoder LSTM to map an input sequence into a fixed-length representation, which is then decoded using single or multiple decoder LSTMs to perform reconstruction of the input sequence or prediction of the future sequence. Then they fine-tune this unsupervised pre-training LSTM to adapt human action recognition tasks. \cite{mahasseni2016action} proposes to train an LSTM that is regularized using the output of another encoder LSTM (RLSTM-g3) grounded on 3D human-skeleton training data, while \cite{action-lstm} proposes a two-step learning method in which 3D features are extracted using CNNs, and then an RNN is trained to classify each sequence by considering the temporal evolution of the learned features. Finally, \cite{LRCN} proposes a class of end-to-end trainable CNN and/or RNN architectures to handle different kinds of inputs. However, the traditional LSTM with matrix multiplications is adopted without exploiting the spatial correlations in the video frames. Directly applying the LSTM models to videos based action recognition is not satisfactory since the spatial correlations and motion dynamics between the frames are not well presented.

To address this, in VideoLSTM \cite{li16videolstm} convolutions are hardwired in the soft-Attention LSTM \cite{SharmaKS15ALSTM}. By stacking another RNN for motion modeling, an enhanced version of attention model is assembled. The input of ConvLSTM, $X_t$, becomes $A_t\odot X_t$ by element-wise multiplying the attention map $A_t$. However, this complex architecture does not bring significant performance improvement. In fact, the performance is highly dependent on the iDT features \cite{iDT}. In other words, VideoLSTM does not characterize the motion dynamics well, even when several attention models are stacked within/on the LSTM.
\section{Models and Algorithm}

\subsection{Revisiting the problem of modeling in RNNs}

\begin{figure*}
  \centering
  \centerline{\includegraphics[width=16.5cm, height=8.5cm]{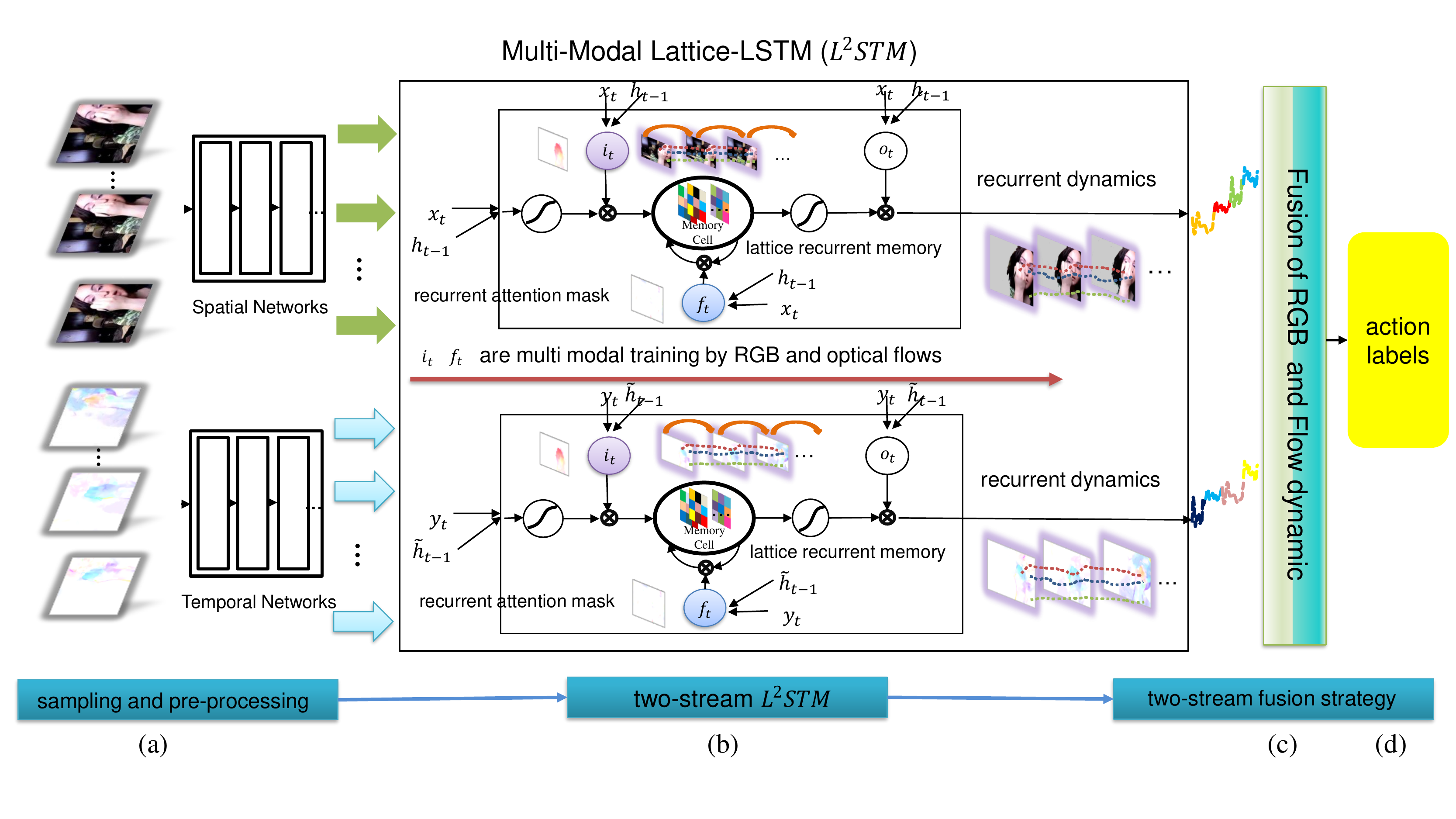}}
  \caption{Illustration of the proposed Lattice-LSTM ($\lstm$). (a) depicts the sampling and pre-processing procedure. A set of video clips (RGB and flow) are passing through CNNs. (b) presents the core functions of the two-stream $\lstm$. Local space-variant superposition operations are applied on the cell memory, enlarging the LSTM's capability to learn the different motion patterns, and the optical flow are fed into the systems simultaneously in order to learn the input and forget gates using multi-modal training. In order to vividly illustrate the system, instead of feature maps, the raw inputs (RGB and corresponding flow) are presented here. (c) is the vectorized learned motion dynamics from $\lstm$ and weighted average fusion. (d) represents the final classification procedure. The whole system is end-to-end trainable.} \label{fig:lstm}
  \vspace{-0.7em}
\end{figure*}
Given a video sequence of length $T$, suppose at each time $t$ with $t = 1, \dots, T$, we can extract at any spatial location $x \in \Omega$ a one-dimensional motion signal $\mathbf{h}_t^x \in \mathbb{R}^L$ of a short length $L$, where $\Omega$ is the 2D spatial domain of the video. To characterize such motion signals, one may use a set of linear filters $\{\Psi_i \in \mathbb{R}^L\}_{i=1}^n$ (e.g., wavelets), and compute their pixel-wise features at time $t$ as $\{ \Psi_i\mathbf{h}_t^x \}_{i=1}^n$.  This is similar to image processing, where a set of pre-defined 2D filters can be used to produce feature maps from an input image via convolution, due to the stationary property of natural images. We also know from the image processing literature that such 2D filters can be learned from a set of training images, so that statistics specific to these training images can be learned to better process images of similar kind \cite{EladSparse}. The learned 2D filters are typically in the format of $5\times 5$ patches. When applied to video processing, this translates to learning a set of filters $\{\Phi_i \in \mathbb{R}^L\}_{i=1}^n$ that are adapted to short-length motion dynamics of training videos. 

However, different from image processing, in many video processing applications, including human action recognition, long-term motion dynamics are important for characterizing different motion patterns. 
Directly learning filters for long-term motion patterns would make both the size $L$ and the number $n$ of filters $\{\Phi_i \in \mathbb{R}^L\}_{i=1}^n$ prohibitively large. This however creates learning difficulties due to increased model complexity that are similar to the difficulties in image processing when learning 2D filters larger than $5\times 5$ patches.

To resolve this issue, one may still learn short-length motion filters and apply them sequentially across the sequence for $t = 1, \dots, T$. However, this scheme cannot take full advantage of modern video processing models such as RNNs/LSTMs. In order to analyze how to resolve this issue when using RNNs/LSTMs, we first investigate how they learn to process video sequences. RNNs use either a hidden layer or a memory cell to learn a time varying set of states which models the underlying dynamics of the input sequence. To better derive and present the limits of RNN, we adopt the general RNN definition without nonlinearities instead of LSTM, since the conclusion from the derivation is the same. We write the recurrent formula of RNNs as follows, where the nonlinearities, control gates, and memory operations are ignored for simplicity:
\begin{eqnarray}\label{eqn:h1}
\begin{aligned}
    H_{t} = W_{H}H_{t-1} + W_{X} X_{t}, \\
\end{aligned}
    \vspace{-1.5em}
\end{eqnarray}
where $\mathbf{X} = \{X_1, X_2, \cdots, X_t\}$ is the input sequence, $X_t$ is an element of the sequence at time $t$, $\mathbf{H} = \{H_1, H_2, \cdots, H_t\}$ is the hidden states, $H_{t}$ is the output of one hidden state at time $t$, and $W_{H}$ and $W_{X}$ are the transition weights. Since the weights are shared across different iterations, the output at the next time step $t+1$ is
\begin{eqnarray}\label{eqn:h2}
\begin{aligned}
     H_{t+1} = W_{H}^{2}H_{t-1} + W_{H}W_{X}X_{t} + W_{X}X_{t+1}. \\
\end{aligned}
\end{eqnarray}
Iteratively, when the time step becomes $t+\tau$, the recurrent formula becomes

\begin{eqnarray}\label{eqn:h3}
\scriptsize
\begin{aligned}
    H_{t+\tau} &= W_{H}^{\tau+1}H_{t-1} + W_{H}^{\tau}W_{X}X_{t} + \cdots + W_{X}X_{t+\tau} \\
    H_{t+\tau} &= \sum_{j=0}^{\tau}W_{H}^{\tau-j}W_{X}X_{t+j} + W_{H}^{\tau+1}H_{t-1} \\
    &=
    \begin{bmatrix}
    W_{H}^{\tau}&W_{H}^{\tau-1}& \dots& 1 \\
    \end{bmatrix}
    \begin{bmatrix}
    W_{X} X_{t} & \\
    W_{X} X_{t+1} & \\
    \vdots  & \\
    W_{X} X_{t+\tau} \\
    \end{bmatrix}
    + W_{H}^{\tau+1}H_{t-1}.
\end{aligned}
\end{eqnarray}
For ease of analysis, we simplify Equ. \ref{eqn:h3} as
\begin{eqnarray}\label{EqnAnalysis1}
H_{t+\tau} = f\left(X_{t:t+\tau}, H_{t-1}\right) , \nonumber
\end{eqnarray}
where $X_{t:t+\tau}$ denotes $\left[ X_t^{\top}, \dots, X_{t+\tau}^{\top} \right]^{\top}$ compactly, $f(\cdot)$ is a linear function to be learned to achieve state transition in LSTM, and $X_{t:t+\tau}$ is a high-dimensional input signal, since it is formed by concatenating all the appearance features from time $t$ to time $t+\tau$. Note in the formula, $W_{H}$ are the only time step related parameters.

It is generally difficult to learn an accurate mapping function $f$ due to its high dimensionality. When $X_{t:t+\tau}$ represents a sequence of video frames, a natural choice is to learn filters that process $X_{t:t+\tau}$ at the local patch level. Indeed, ConvLSTM \cite{xingjian2015convolutional} learns the same set of patch-level filters for different spatial locations of videos. The form of Equ. \ref{eqn:h3} (i.e., polynomials of the basic filters $W_H$) reduces the search space of high-dimensional filters to be learned, and removes some possible patterns that the learned filters should characterize. We would like to increase the model capacity to better characterize complex motion patterns. In machine learning, a traditional way to learn a function with a higher model capacity is to partition the feature space into local cells and learn separate mappings in each local cell. In this work, we follow this idea and propose to partition the high-dimensional feature space into local cells based on their positions in the spatial domain. Thus the difficulty of learning mapping functions in high-dimensional space is reduced to learning mappings for each pixel separately.

\subsection{Lattice-LSTM ($\lstm$)}
In this section, we propose Lattice-LSTM ($\lstm$) to solve the issues presented in the above analysis and derivation. In order to enhance the modeling of motion dynamics, the newly generated cell memory $\widetilde{C}_t$ at time $t$ of $\lstm$ is formulated as

\begin{eqnarray}\label{eqn:memorycell}
\begin{aligned}
\widetilde{C}_t &= tanh(W_{xc} \ast X_t + W_{hc}\textcircled {\small{$\textsl{L}$}}H_{t-1}),\\
\end{aligned}
\end{eqnarray}
where $W_{xc}$ and $W_{hc}$ are the distinct weights for the input and hidden state, $\ast$ denotes the conventional convolution operation, and $\textcircled{\footnotesize{$\textsl{L}$}}$ denotes a local superposition summation.. The local superposition is defined as follows:
\begin{eqnarray}\label{eqn:lattice}
\footnotesize
\begin{aligned}
&W_{hc}\textcircled{\scriptsize{$\textsl{L}$}}H_{t-1} \\
&= \forall_{k,i,j}(\sum_{l,m,n}W_{hc}(l,i,j,k, m,n)H_t(l, i+m-1, j+n-1)),
\end{aligned}
\end{eqnarray}
where $(i,j)$ indicates the position, $m$ and $n$ index position in the kernel, and $l$ and $k$ index the input and output feature maps. Intuitively, this is performing a local filtering operation with different filters at different spatial locations.

These local filters, which are initialized independently, are applied on each memory cell pixel in order to learn the different temporal patterns at different pixels. According to Equ. \ref{eqn:h3}, we only apply the local superposition on the hidden transition of cell memory within the recurrence and the other linear combinations are convolutional. Therefore, $W_{hc}$ has a larger capacity than $W_{xc}$. Since the memory cell stores the information obtained over time, the local superposition will enhance the capability of preserving dynamics in the memory cell and model spatial nonhomogeneity in the long term dependency. Here the term 'lattice' is used to vividly indicate the local superposition.  At the same time, the input and forget gates which control the memory cell are
\begin{eqnarray}\label{eqn:gates}
\begin{aligned}
i_t &= \sigma(W_{xi}\ast X_t + W_{hi}\ast H_{t-1})\\
f_t &= \sigma(W_{xf}\ast X_t + W_{hf}\ast H_{t-1}),\\
\end{aligned}
\end{eqnarray}
where $W_{xi}, W_{hi}$ and $W_{xf}, W_{hf}$ are distinct weights for the input and forget gates respectively. The updated memory cell $C_t$ can be represented as
\begin{eqnarray}\label{eqn:memory}
\begin{aligned}
C_t &= f_t\circ C_{t-1} + i_t\circ \widetilde{C}_t,\\
\end{aligned}
\end{eqnarray}
where $C_{t-1}$ is the previous memory cell and `$\circ$' denotes the Hadamard product. The input gates and forget gates together determine the amount of dynamic information entering/leaving the memory cell. Since the input gates and forget gates are important for controlling the memory cell, during the learning of $\{W_{xi}, W_{hi}, W_{xf}, W_{hf}\}$, we use a multi-modal learning procedure. RGB frames of the video and corresponding optical flow are fed into the system simultaneously to learn the shared input gates and the shared forget gates for the two-stream architecture. Therefore the gradient update for the input gate and forget gate should be
\begin{eqnarray}\label{eqn:convLSTM}
\begin{aligned}
W_{\cdot i} := W_{\cdot i} - \eta(\bigtriangledown Q^{RGB}(W_{\cdot i})+ \bigtriangledown Q^{Flow}(W_{\cdot i}))\\
W_{\cdot f} := W_{\cdot f} - \eta(\bigtriangledown Q(^{RGB}W_{\cdot f}) + \bigtriangledown Q^{Flow}(W_{\cdot f})),
\end{aligned}
\end{eqnarray}
where $Q^{RGB}(\cdot)$ and $Q^{Flow}(\cdot)$ are the estimated loss functions passing to the memory cell from the RGB inputs and corresponding flow, $\cdot$ denotes the input and hidden state and $\eta$ is the learning rate. With multi-modal learning, the controlled gates can well adjust the information entering/leaving the cell memory. Finally, the output of one iteration of $\lstm$ is
\begin{eqnarray}\label{eqn:convLSTM}
\begin{aligned}
o_t &= \sigma(W_{xo}\ast X_t + W_{ho}\ast H_{t-1})\\
H_t &= o_t\circ tanh(C_t),
\end{aligned}
\end{eqnarray}
where $W_{xo}$ and $W_{ho}$ are distinct weights for the output gate.

Because we consider the spatial variability in the temporal modeling, the representations of motion dynamics have been greatly enhanced due to the local superposition. Therefore, more motion patterns existing in videos can be depicted and captured. Meanwhile, the multi-modal-trained control gates provide motion masks for the memory cell in order to generate distinguishing representations. Batch normalization \cite{ioffe2015batch} is applied after each spatial convolutional operation.

\begin{figure}[ht]
  \centering
  \centerline{\includegraphics[width=8cm]{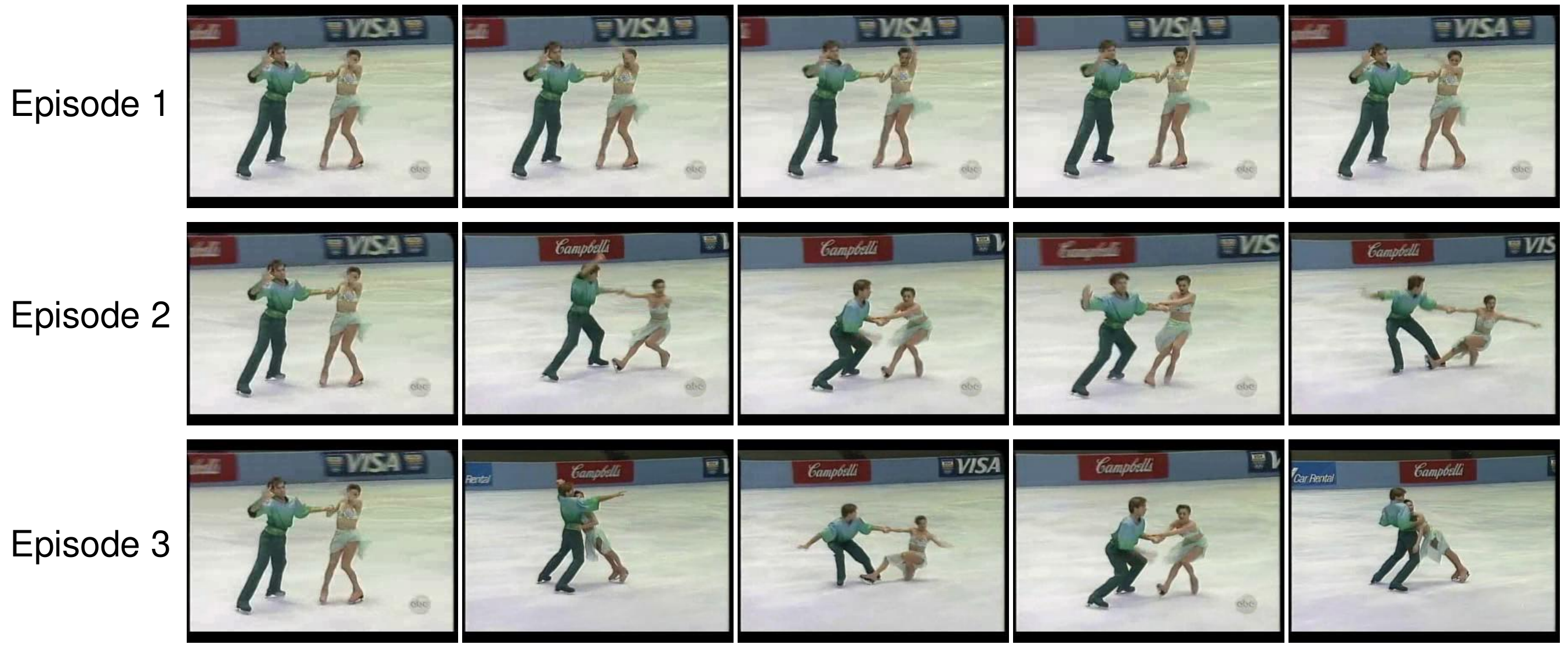}}
  \caption{Frames extracted from an ice dancing video. Sampling using different strides makes different episodes. Episode 1 is almost stationary, Episode 2 becomes time varying and Episode 3 has dramatic change. } \label{fig:episode}
  \vspace{-0.9em}
\end{figure}

\subsection{Long-Short Term Sampling}\label{sec:sample}
Due to computational limitations, we cannot feed all sequential inputs into the system and train them, so sampling is necessary for training the videos end-to-end. As shown in Fig. \ref{fig:episode}, even from the same video sequence, different sampling methods can generate different episodes. In this section, we implement a new sampling strategy in order to make the recurrent networks learn more reasonable `long' and `short' representations.

Inspired by \cite{fstcn}, we apply temporal data augmentation for training the network. Suppose we are given a video sequence $\mathbf{V} \in \mathbb{R}^{m_x\times m_y\times m_t}$, where ($m_x$, $m_y$) are the spatial dimensions of the video and $m_t$ is the number of frames in the video sequence. Given this video sequence and a patch size of $l_x\times l_y$ (such that $l_x < m_x, l_y < m_y$, $l_t < m_t$), we can sample a set of $n$ video clips $S = \{\mathbf{V}_{clip}^{(1)}, \mathbf{V}_{clip}^{(2)}, \ldots, \mathbf{V}_{clip}^{(n)}\}$ which constitute a single forward pass through the unrolled model (with $n$ steps) as follows. To sample a video clip $\mathbf{V}_{clip} \in \mathbb{R}^{l_x\times l_y\times l_t}$, we sample $l_x\times l_y$ crops of $l_t$ frames with a temporal stride of $s_t$, and we use a spacing of $s_s$ frames between each video clip in the sequence to form our set $S$. Note that for each individual sequence, we sample a new crop center, a new stride $s_s$ such that $0\leq s_s\leq s_{max}$, where $s_{max}$ is the maximum possible stride, and a new stride $s_t$ such that $0\leq s_t\leq s_s$.

Depending on our sampling of the parameters, we can select a sequence of clips $S$ covering a very short portion of the video or a sequence of clips $S$ covering a long portion of the video. In other words, this kind of sampling can provide `short'-term clips, which cover consecutive frames ($s_t=1$ and $s_s$ is the minimum stride) and in which the movement is stationary, and `long'-term clips, which cover the whole video ($s_t = s_s$ and $s_s=s_{max}$) and in which the movement changes a lot. All LSTM models can benefit from this sampling formulation. We see about a $2\%$ gain compared with the traditional sampling that uses a fixed stride and randomly extracted clips, and we have adopted our temporal data augmentation in all of our experiments.

\subsection{Learning Architecture}
The whole architecture is presented in Fig. \ref{fig:lstm}. In order to illustrate the concept clearly, instead of using feature maps from a CNN, in the figure, we use raw RGB and optical flow images to illustrate the memory cell, input, forget and output gates. Local superposition operations are applied on the cell memory to form the lattice recurrent memory, enhancing the LSTM's ability to learn complex motion patterns and addressing the long-term dependency issue.

As illustrated in Fig.~\ref{fig:lstm}, we adopt the two-stream framework in which the optical flow is fed as an additional modality to further compensate and improve the prediction of RGB inputs. Note that video clips, each composed of several frames sampled from a video, are fed through the pre-trained spatial and temporal networks at each time step to extract the high-level feature maps. Compared to traditional two-stream frameworks which train each stream independently (on the corresponding input modality), combining the scores together at the end, our model feeds in the RGB and flow information simultaneously in order to learn the shared input gates and shared forget gates that form the recurrent attention mask for the memory cell. Note that only the input and forget gates are shared between the RGB and flow, while the others are learned independently.

This multi-modal learning procedure allows the input and forget gates to leverage both appearance and dynamic information. The learned recurrent attention mask of the input and forget gates, which can further regularize enhanced dynamics from the memory cell, is formed to control the entering/leaving dynamics from the memory cell. After regularization of the input and forget gates, the output should be motion features which can capture very complex dynamics. Each memory cell and output gate is learned independently of the other stream in order to optimize the features to each modality. The final prediction is the weighted average of the two outputs from the sequential RGB and flow. The video sequence is recognized as the action category corresponding to the highest probability.

\section{Experiments}\label{experiments}
Experiments are mainly conducted on two action recognition benchmark datasets, namely UCF-101 \cite{UCF-101} and HMDB-51 \cite{HMDB-51}, which are currently the largest and most challenging annotated action recognition datasets.

\textbf{UCF-101 } \cite{UCF-101} is composed of realistic web videos with various camera motions and illuminations. It has more than 13K videos, with an average length of 180 frames per video categorized into 101 actions, ranging from daily life activities to unusual sports.

\textbf{HMDB-51 } \cite{HMDB-51} has a total of 6766 videos organized as 51 distinct action categories. This dataset is more challenging than others because it contains scenes with more complex backgrounds. It has similar scenes in different categories, and it has a small number of training videos.

Both UCF-101 and HMDB-51 have three split settings to separate the dataset into training and testing videos. We report the mean classification accuracy over these splits.

\subsection{Implementation Details}
We choose part of VGG16 \cite{VGG16}, which consists of 13 convolutional layers, as our CNN feature extractor for the RGB and optical flow images. The spatial and temporal networks are pre-trained on ImageNet \cite{imagenet} and finetuned on the corresponding human action recognition datasets. $\lstm$ is trained from scratch. The input of the optical flow network is stacked optical flow images, each of which has two channels (the horizontal and vertical flow fields). The optical flow \cite{optflow} is computed from every two adjacent frames. We linearly rescale the values of the flow fields to [0, 255]. We feed in a stack of five RGB frames to the spatial network and a stack of five corresponding flow frames to the temporal network. Since the pre-trained VGG16 network uses inputs of different dimensions, we re-train the first layer of both networks. Then the features from the last pooling layer (e.g. pool5) of each VGG16 network are fed into $\lstm$. 

Our implementation contains 8 unrolled time steps. Therefore, eight video clips, composed of 5 frames each, are regularly extracted from the videos with the random strides (as described in the long-short term sampling) for each $\lstm$ training step using back-propagation through time (BPTT). Within $\lstm$, all convolutional kernels and superposition kernels are $3\times 3$. We use mini-batch stochastic gradient descent (SGD) to train the network with an initial learning rate of 0.001, a momentum of 0.9, and a weight decay $5\times10^{-4}$. We decay the learning rate when it is saturated and adopt two-step training. That is, we first fix the CNN weights and train the $\lstm$ only. When the accuracy of the $\lstm$ on the validation videos stops increasing, we fine-tune the whole network using a smaller learning rate. For data augmentation, we use color jittering, as described  in \cite{AlexNet}, horizontal flipping, and random cropping for the spatial network and horizontal flipping and random cropping for the temporal network. When testing, more video clips are fed, and we accumulate different probabilities from different length sequences with different strides in order to combine `long' and `short' predictions.
 
Video clips sampled from videos are fed into the system and scores are extracted from different time steps of $\lstm$. The final scores are the combination of these scores $[S_1, S_2, S_3,\cdots S_t]$, where $S$ is the output score and t is the time step with different strides. The final scores should be the average of these outputs with different strides, that is, $S = mean(S_1, S_2, S_3,\cdots, S_t)$. Real-time action recognition can be achieved when the video is streaming sequentially. The video streaming can be fed into the system sequentially, and the prediction accuracy can be accumulated when more clips are fed in.
\begin{figure}
  \centering
  \centerline{\includegraphics[width=8.5cm]{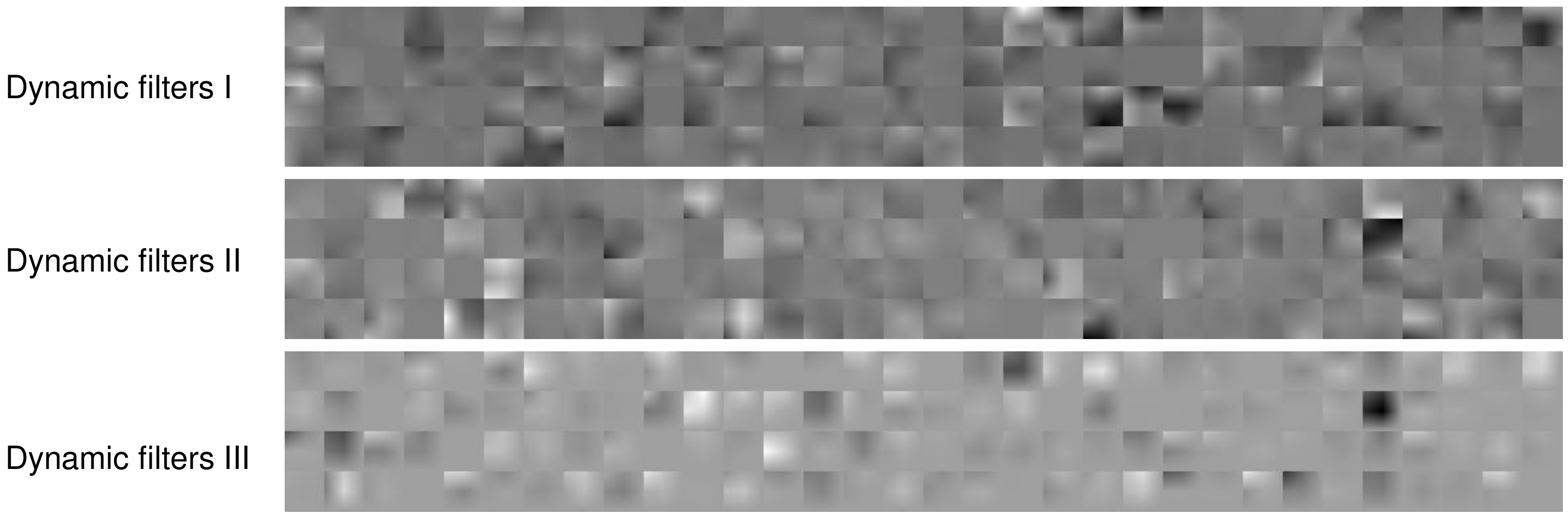}}
  \caption{The visualization of several leaned kernels in the local superposition operator. For each filter, we visualize the kernels of 128 channels. Note that we rescale the $3 \times 3$ filters to make the visualization more pleasing.} \label{fig:filters}
\end{figure}

\subsection{Visualization}
Several learned local superposition kernels are shown in Fig. \ref{fig:filters}. Due to the page limit, we present three sets of learned position-varying filters, each of which has 128 channels. We can clearly see that these sets of filters learn different properties, further verifying our previous assumptions; different superposition may represent the different semantic meaning during actions.

\begin{figure}
  \centering
  \centerline{\includegraphics[width=8cm]{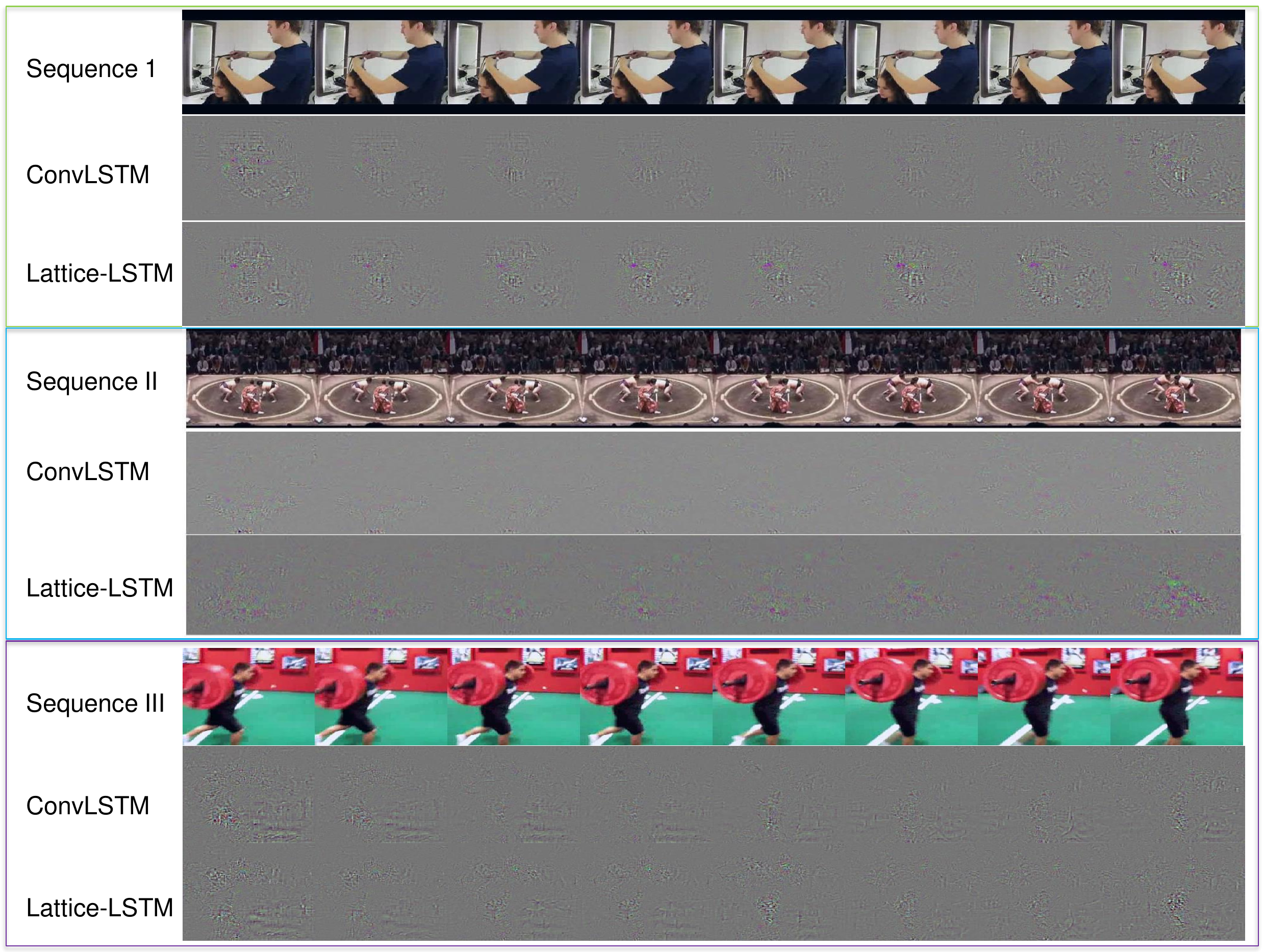}}
  \caption{Saliency map comparison. From top to bottom is the original video sequence, ConvLSTM and our $\lstm$. As shown in the figure, $\lstm$ can accurately detect the interactive and moving points, while ConvLSTM can not. This visualization looks better in color and high resolution. } \label{fig:saliency}
\end{figure}

To visually verify the effect of $\lstm$, we use BPTT to visualize saliency regions for specific video sequences, i.e., back-propagating the neuron of that action category in the classifier layer to the input image domain and plotting the corresponding gradient. As shown in Fig. \ref{fig:saliency}, our $\lstm$ can catch the most dynamic regions, while ConvLSTM cannot for the sake of treating all movements the same. For instance, in Sequence I, $\lstm$ can provide accurate motion region detection, even when the movement is very small (scissors moving around a head), while ConvLSTM can not. What is more, in Sequence III, $\lstm$ can detect a leg movement and a lifting movement, while ConvLSTM only focuses on the leg movement. Owing to the local superposition operation on the cell memory, our proposed model can provide more dynamics through recurrence. At the same time, thanks to the multi-modal learning, the input and forget gates can control the dynamic entering/leaving well, so useless dynamics are not present in the salient regions.

\subsection{Effect on videos with complex movements}
We specifically evaluated our model on videos with complex movements. Besides its 101 categories, UCF-101 has coarse definitions which divide the videos into human and object interaction, human and human interaction and sports. We evaluate the performance on these coarse categories, and as reported in Table. \ref{t:inter}. The upper part of the table lists the performance on the coarse category. Compared with ConvLSTM (note here the only difference between ConvLSTM and $\lstm$ is the network architecture), our model performs much better on all coarse categories, particularly on human and object interaction $\uparrow 10.7\%$, human and human interaction $\uparrow 6.3\%$ and body motion only $\uparrow 0.5\%$. Since our model can well handle the complex movements while ConvLSTM may not, it performs much better on the human interacting actions. On the other hand, they perform similarly on the body motion only actions since the movements are simple in this category, which verifies our hypothesis. The lower part of the table shows the performance for specific categories. To provide an overview, the sample frames from each coarse category can be found from the dataset website. We can see from the table that the more complex the videos are, the greater the percentage improvement there is. In the pizza tossing category in UCF-101, the movements of the person and pizza are complex and fast, and we see a $51.5\%$ improvement.

\begin{table}[htb] \caption{Performance evaluation on videos of split 1 of UCF-101 with complex movements} \label{t:inter}
\footnotesize
\centering
\begin{tabular}{c|c|c|c} \hline
Data Types & ConvLSTM & $\lstm$ &  Gain \\
\hline
Human-Object Interaction &76.0 &86.7  & \textbf{$\uparrow$ 10.7}   \\
Human-Human Interaction  &89.1 &95.4  & \textbf{$\uparrow$ 6.3} \\
Body Motion Only                   &88.1 &88.6  & \textbf{$\uparrow$ 0.5}  \\
\hline
\hline
Pizza Tossing    &21.2 &72.7 & \textbf{$\uparrow$ 51.5}   \\
Mixing Batter    &55.6 &86.7 & \textbf{$\uparrow$ 31.1}   \\
Playing Dhol     &81.6 &100 & \textbf{$\uparrow$ 18.4}    \\
Salsa Spins      &86.0 &100  & \textbf{$\uparrow$ 4.0 }   \\
Ice Dancing       &97.8 &100  & \textbf{$\uparrow$ 2.2}    \\
\hline
\end{tabular}
\end{table}

\subsection{Comparison with LSTM-like architecture}
In Table. \ref{t:comp}, we list the performance of all LSTM variants and our main component for action recognition of spatial and temporal networks on split 1 of UCF-101. Our proposed method significantly outperforms the variants on each network. From the component analysis, we find that the local superposition has a more significant impact on the temporal networks than spatial networks, since more useful features are extracted from flow using $\lstm$. Additionally, the improvement of the shared control gates on the spatial networks is higher than on the temporal networks, indicating that spatial networks benefit more from multi-modal training.

\begin{table}\caption{$\lstm$ components analysis on split 1 of UCF-101} \label{t:comp}
\footnotesize
\centering
\begin{tabular}{c|c|c} \hline
Method & Spatial Networks & Temporal Networks\\
\hline
LSTM \cite{li16videolstm} &77.5&78.3\\

ALSTM \cite{li16videolstm} &77.0&79.5\\

VideoLSTM \cite{li16videolstm} &79.6 & 82.1\\
\hline
\hline
ConvLSTM  &77.6 &79.1 \\
+long-short term sampling &80.4 &82.3 \\
+locally superpostion     &82.2 &86.9 \\
+shared controlling gates &83.2 &87.4 \\ 
\hline
\end{tabular}
\end{table}

\begingroup
\setlength{\tabcolsep}{2.5pt}
\renewcommand{\arraystretch}{1.3}
\begin{table}\caption{State-of-the-art comparison with LSTM-like architectures} \label{t:lstmalgo}
\scriptsize
\centering
\begin{tabular}{c|c|c|c|c|c|c|c|c} \hline
\multirow{2}{*}{Method} &  \multicolumn{2}{c}{Pre-training} &  \multicolumn{2}{c}{Networks} &\multicolumn{2}{c}{\multirow{2}{*}{UCF101 HMDB51}} \\
& \multicolumn{1}{c}{ImageNet} &\multicolumn{1}{c}{1M Sports} &\multicolumn{1}{c}{VGG-M} &\multicolumn{1}{c}{VGG16} \\
\hline
 LRCN\cite{LRCN}            &\multicolumn{1}{c}{\checkmark}&\multicolumn{1}{c}{-} &\multicolumn{1}{c}{\checkmark}&\multicolumn{1}{c}{-} &\multicolumn{1}{c}{82.9} &\multicolumn{1}{c}{-} \\
 TwoLSTM\cite{two-lstm}      &\multicolumn{1}{c}{\checkmark} &\multicolumn{1}{c}{\checkmark} &\multicolumn{1}{c}{-}&\multicolumn{1}{c}{\checkmark} &\multicolumn{1}{c}{88.3}&\multicolumn{1}{c}{-}\\
 ALSTM\cite{SharmaKS15ALSTM}   &\multicolumn{1}{c}{\checkmark}&\multicolumn{1}{c}{-} &\multicolumn{1}{c}{-}&\multicolumn{1}{c}{\checkmark} &\multicolumn{1}{c}{77} &\multicolumn{1}{c}{41.3}\\
 RLSTM-g3\cite{mahasseni2016action} &\multicolumn{1}{c}{\checkmark}&\multicolumn{1}{c}{-} &\multicolumn{1}{c}{-}&\multicolumn{1}{c}{\checkmark} &\multicolumn{1}{c}{86.9}&\multicolumn{1}{c}{55.3} \\
 UnsuperLSTM\cite{Srivastava_unsupervisedlearning}  &\multicolumn{1}{c}{\checkmark} &\multicolumn{1}{c}{\checkmark} &\multicolumn{1}{c}{-}&\multicolumn{1}{c}{\checkmark} &\multicolumn{1}{c}{84.3} &\multicolumn{1}{c}{44.0}\\
 VideoLSTM\cite{li16videolstm}  &\multicolumn{1}{c}{\checkmark}&\multicolumn{1}{c}{-} &\multicolumn{1}{c}{-}&\multicolumn{1}{c}{\checkmark} &\multicolumn{1}{c}{89.2} &\multicolumn{1}{c}{56.4}\\
\hline
\hline
$\lstm$   &\multicolumn{1}{c}{\checkmark}&\multicolumn{1}{c}{-} &\multicolumn{1}{c}{-}&\multicolumn{1}{c}{\checkmark} &\multicolumn{1}{c}{\textbf{93.6}}&\multicolumn{1}{c}{\textbf{66.2}}\\
\hline
\end{tabular}
\vspace{-1em}
\end{table}
\endgroup

In Table. \ref{t:lstmalgo}, we list all state-of-the-art methods which use an LSTM-like architecture for action recognition tasks. In order to present the comparison completely and clearly, we elaborate on some factors, such as, pre-training type and network architecture. From the table, our proposed method clearly performs the best amongst the LSTM-like architectures, even though we do not have 1M sports pre-training.

\subsection{Comparison with the-state-of-the-art}
The temporal information can be well modeled using $\lstm$. In addition to the LSTM-like comparison in Table \ref{t:lstmalgo}, another state-of-the-art algorithm comparison is presented in Table \ref{t:results}, where we compare our method with traditional approaches such as improved trajectories (iDTs) \cite{iDT} and deep learning architectures such as two-stream networks \cite{TwoStream}, factorized spatio-temporal convolutional networks ($\FSTCN$) \cite{fstcn} (which use VGG-M \cite{Zeiler14} as a deep feature extractor), 3D convolutional networks (C3D) \cite{C3D}, trajectory-pooled deep convolutional descriptors (TDD) \cite{tdd}, and spatio-temporal fusion CNNs \cite{Feichtenhofer16fusion}, which use a VGG16 deep architecture on the UCF-101\cite{UCF-101} and HMDB-51\cite{HMDB-51} datasets. We also include some complex network-based methods: ST-ResNet \cite{Feichtenhofer16resinet} and TSN \cite{linmin16segment}. Compared with the traditional hand-crafted methods and deep learning methods, our proposed $\lstm$ achieves better performance.

\begin{table}[htb] \caption{Mean accuracy on the UCF-101 and HMDB-51} \label{t:results}
\small
\centering
\begin{threeparttable}
\begin{tabular}{|c|c|c|c|}
\hline
Model& Method & UCF-101 & HMDB-51     \\
\hline
\multirow{4}{*}{Traditional}& iDT + FV \cite{iDT} &85.9 &57.2 \\
& iDT + HSV \cite{iDTHSV} &87.9 &61.1 \\
& VideoDarwin \cite{video_evolution} &- &63.7\\
& MPR \cite{MPR} &- &65.5\\
\hline
\hline
\multirow{3}{*}{Deep}& EMV-CNN \cite{zhang16mv} &86.4 &- \\
&Two Stream \cite{TwoStream} &88.0 &59.4\\
&$F_{ST}CN$ \cite{fstcn} &88.1 &59.1\\
\hline
\hline
\multirow{4}{*}{Very Deep}&C3D (3 nets) \cite{C3D} &85.2 &-\\
&VideoLSTM\cite{li16videolstm} &89.2 &56.4 \\
&TDD+FV \cite{tdd} &90.3 &63.2\\
&Fusion \cite{Feichtenhofer16fusion} &92.5 &65.4 \\
\hline
\multirow{1}{*}{Ours}& $\lstm$  & \textbf{93.6} & \textbf{66.2} \\
\hline
\hline
\multirow{2}{*}{Complex \tnote{*}}&ST-ResNet \cite{Feichtenhofer16resinet}   &93.4 &66.4 \\
&TSN \cite{linmin16segment} &94 &68.5\\
\hline
\end{tabular}
\begin{tablenotes}
\scriptsize
\item[*] Take advantages of complex deep architecture; their spatial and temporal networks (baseline) are about $1\%$ to $2\%$ better than ours.
\end{tablenotes}
\end{threeparttable}
\end{table}

\section{Conclusion}
In order to solve existing problems in action recognition, $\lstm$ is proposed. This method applies local spatially varying superposition operations to the memory cell, and also uses control gates (input and forget) trained using RGB and flow. We also introduced a novel long-short term sampling method which boosts the performance even further. A huge gain is observed compared with other LSTM-like architectures for action recognition tasks. This model achieves state-of-the-art performance even compared to more complex networks. High level feature maps from CNNs contain semantic meanings, which can be modeled using different filters across time. In this paper, we learn these filters from space-variant superposition. An extension would be to adaptively group these semantic meanings based on videos, not learning them on each location. We hope this work can shed light on LSTM architectures for video analysis.

\noindent \textbf{Acknowledgements:} We gratefully acknowledge the support of the Hong Kong University of Science and Technology, MURI (1186514-1-TBCJE), ONR ( 1196026-1-TDVWE);  NISSAN (1188371-1-UDARQ. This work also used the Extreme Science and Engineering Discovery Environment (XSEDE), which is supported by National Science Foundation grant number ACI-1548562.

{
\bibliographystyle{ieee}
\bibliography{LSTM}
}

\end{document}